\documentclass{article}

\usepackage{microtype}
\usepackage{graphicx}
\usepackage{subfigure}
\usepackage{booktabs} 
\usepackage{subcaption}
\usepackage{adjustbox}

\usepackage{hyperref}

\usepackage[accepted]{mlsys2025}

\mlsystitlerunning{BambooKG: A Neurobiologically-inspired Frequency-Weight Knowledge Graph}

\begin{document}

\twocolumn[
\mlsystitle{BambooKG: A Neurobiologically-inspired Frequency-Weight Knowledge Graph}



\mlsyssetsymbol{equal}{*}

\begin{mlsysauthorlist}
\mlsysauthor{Vanya Arikutharam}{equal,ulla}
\mlsysauthor{Arkadiy Ukolov}{equal,ulla}
\end{mlsysauthorlist}

\mlsysaffiliation{ulla}{Ulla Technology (OWM Group), London, England, UK}

\mlsyscorrespondingauthor{Arkadiy Ukolov}{arkady@openweather.co.uk}

\mlsyskeywords{Knowledge Graph, triplets, memory recall, embeddings}

\vskip 0.3in

\begin{abstract}
Retrieval‑Augmented Generation (RAG) allows LLMs to access external knowledge, reducing hallucinations and ageing‑data issues. However, it treats retrieved chunks independently and struggles with multi‑hop or relational reasoning, especially across documents \cite{tang2024multihop}. Knowledge Graphs (KG) enhance this by capturing the relationships between entities using triplets, enabling structured, multi‑chunk reasoning; however, these tend to miss information that fails to conform to the triplet structure \cite{peng2024graphrag_survey}. We introduce BambooKG, a knowledge graph with frequency‑based weights on non-triplet edges which reflect link strength, drawing on the Hebbian principle of “fire together, wire together” \cite{hebb1949organization}. This decreases information loss and results in improved performance on single- and multi-hop reasoning, outperforming the existing solutions.
\end{abstract}
]




\printAffiliationsAndNotice{\mlsysEqualContribution} 

\section{Introduction}
\label{introduction}

Human cognition exhibits a remarkable ability to form and retain associations between related experiences, a process underpinned by Spike-Timing Dependent Plasticity (STDP) and the Hebbian principle of “neurons that fire together, wire together” \cite{hebb1949organization, caporale2008stdp, bi1998synaptic}. Through repeated co-activation, synaptic pathways strengthen over time, enabling associative memory, the dynamic capacity to recall concepts through partial cues rather than exact matches. This biological principle has long inspired computational models of learning and retrieval, from Hopfield networks \cite{hopfield1982neural} to modern energy-based and graph-structured memory and LLM architectures \cite{bartunov2020meta, kosowski2025dragon}.

In this work, we present \textbf{BambooKG: Biologically-inspired Associative Memory Based On Overlaps KG}. BambooKG is a neurobiologically-motivated framework for long-term memory retention in knowledge graphs. It introduces a frequency-weighted associative mechanism, where the repeated co-occurrence of non-triplet node pairs - derived from shared semantic or contextual “chunks” - incrementally strengthens the edge weights that connect them. 

Unlike static embedding spaces, BambooKG evolves as a function of activation history. Each tagging event triggered by the addition of new information contributes to the edge frequency distribution, effectively encoding temporal salience and contextual relevance. This enables the system to “associate” and connect new information with existing knowledge, thereby enabling effective retrieval during recall.


\begin{figure*}[!t]
  \centering
  \begin{subfigure}[]
    \centering
    \includegraphics[width=\columnwidth]{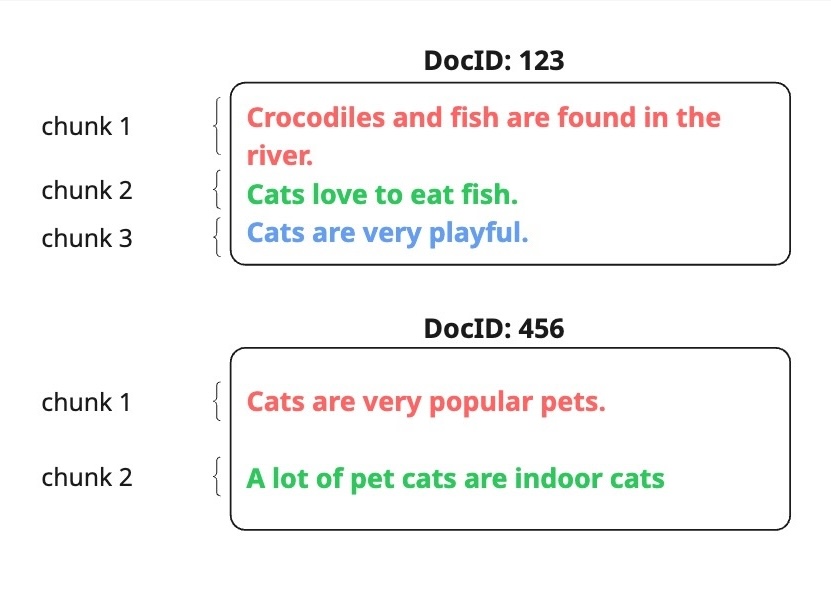}
    \label{fig:chunking_stage}
  \end{subfigure}
  \hfill
  \begin{subfigure}[]
    \centering
    \includegraphics[width=\columnwidth]{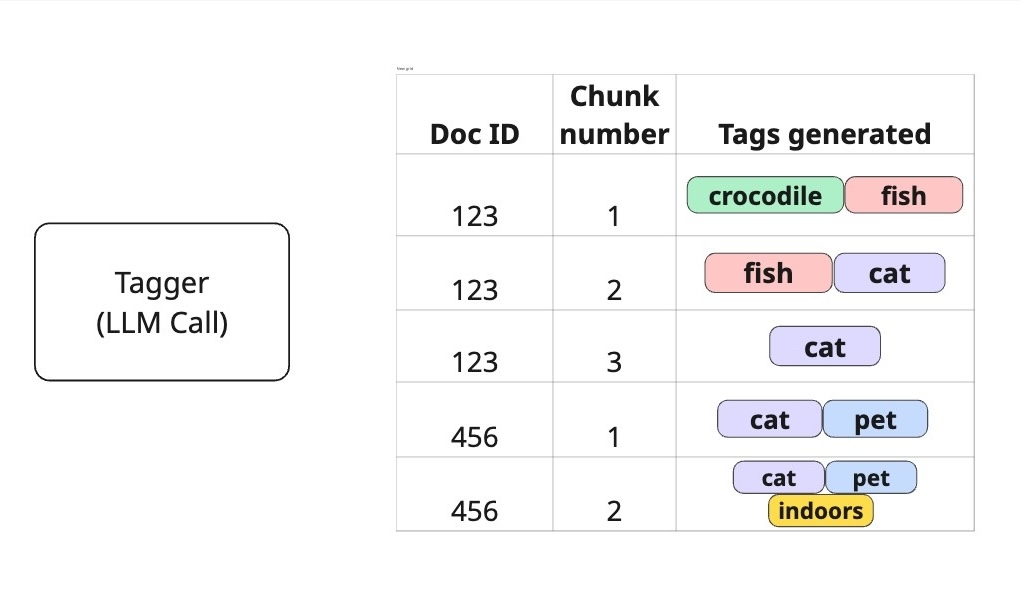}
    \label{fig:tag_generation}
  \end{subfigure}

  \vspace{1em} 

  \begin{subfigure}[]
    \centering
    \includegraphics[width=0.98\textwidth]{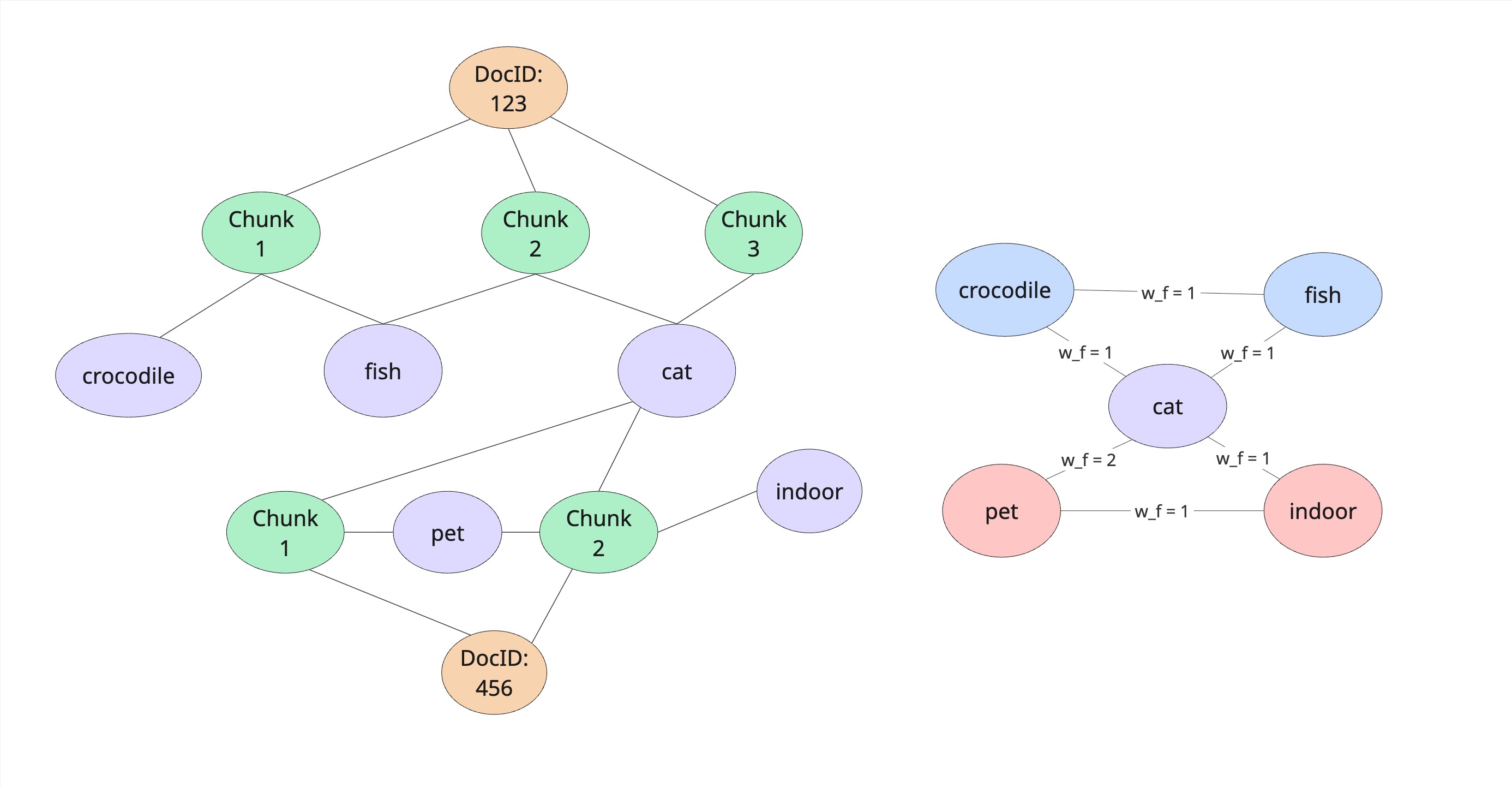}
    \label{fig:graph_creation_stage}
  \end{subfigure}

  \caption{Overall memory pipeline: (a) Chunking, (b) Tag Generation, and (c) Knowledge Graph Creation stages (\textit{left} - tag to chunk mapping KG and \textit{right} - tag frequency weight KG)}
  \label{fig:memory_pipeline_all}
\end{figure*}

\section{Existing Models}
\subsection{RAG}
RAG is a standard document retrieval and reasoning approach that involves simple cosine similarity search over the embedded documents. It is widely applied in domains such as healthcare and enterprise QA. \cite{yang2025raghealth}. Despite its strengths, RAG struggles with multi-hop reasoning because it treats retrieved chunks independently \cite{tang2024multihop}. Newer pipelines, such as Chain-of-RAG, achieve state-of-the-art results on the KILT benchmark, showing over 10-point EM score improvements for multi-hop QA \cite{wang2025chain}. However, stepwise retrieval introduces higher computational overhead and longer inference times. Its performance can degrade if intermediate retrieval steps accumulate errors or irrelevant passages \cite{wang2025corag}. Multi-agent optimisation strategies also improve QA F1 scores by jointly training retrieval, filtering, and generation modules. Nevertheless, it significantly increases training complexity, requires careful reward design, and can be unstable \cite{chen2025mmoarag}.

\subsection{OpenIE}
OpenIE systems extract structured triples (subject–relation–object) directly from text, allowing flexible knowledge graph construction without predefined schemas \cite{etzioni2015openie}. These models underpin many downstream reasoning frameworks, including GraphRAG. OpenIE captures fine-grained relations across sentences, but suffers from low precision in noisy or domain-specific corpora. Early benchmarks showed competitive recall but lower F1 compared to supervised relation extractors (e.g., F1 scores ~50–60\% in heterogeneous corpora) \cite{etzioni2011openie}, limiting its standalone utility.

\subsection{GraphRAG}
GraphRAG is an attempt at a combination between RAG and OpenIE. It builds knowledge graphs from retrieved passages, allowing entity disambiguation, relational reasoning, and multi-hop synthesis \cite{edge2024graphrag}. Recent work integrating causal graphs has shown up to 10\% absolute accuracy improvements in medical QA tasks compared to standard GraphRAG \cite{luo2025causalgraphrag}. However, performance gains depend on graph construction quality; noisy relation extraction or sparse knowledge domains reduce effectiveness. GraphRAG introduces higher computational overhead, but benchmarks suggest it outperforms standard RAG in complex multi-hop QA where relational inference is required. \cite{Han2025RAGvsGraphRAG}

\subsection{KGGen}
Another knowledge graph model that we compare BambooKG against is KGGen. The fundamental idea lies in using multiple LLM calls to create a knowledge graph. The process consists of several stages: extraction of entities, extraction of relations between them, aggregation and clusterisation. Each step involves one or several LLM calls. The benefit of this approach is the increased connectivity between articles \cite{MoYuKazdan2025_KGGen}.

\begin{figure*}[!t]
  \centering
  \includegraphics[width=\textwidth]{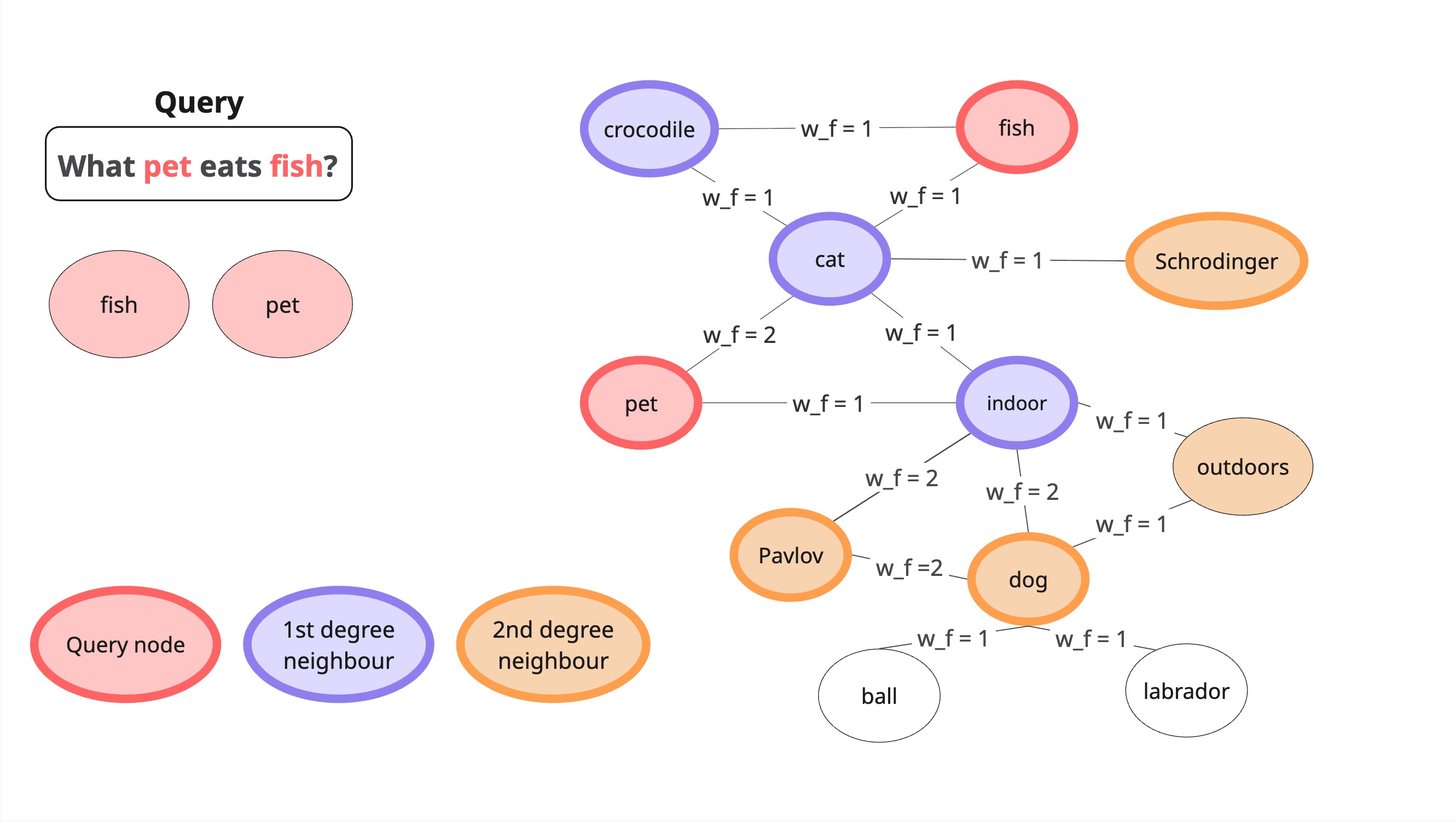}
  \caption{Query Subgraph Generation. For illustrative purposes we take the top 2 first degree neighbours and the top 2 second degree neighbour for each query node. Note: some connections are not shown to ensure the readability of the diagram}
  \label{fig:query_subgraph_generation}
\end{figure*}

\begin{figure*}[!t]
  \centering
  
  \begin{subfigure}[]
    \centering
    \includegraphics[width=\textwidth]{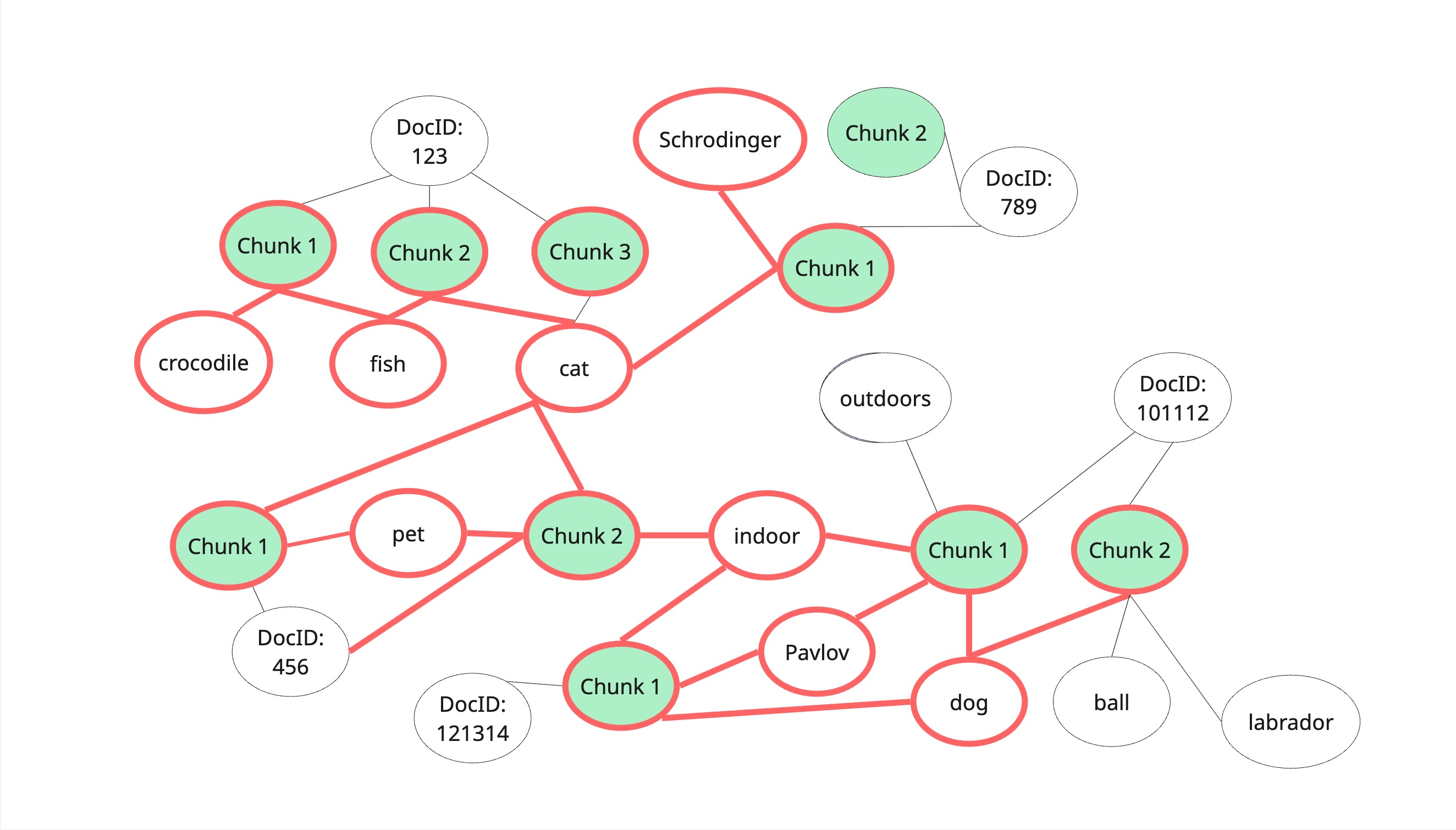}
    \label{fig:chunk_extraction}
  \end{subfigure}

  \vspace{1em} 

  \begin{subfigure}[]
    \centering
    \includegraphics[width=\textwidth]{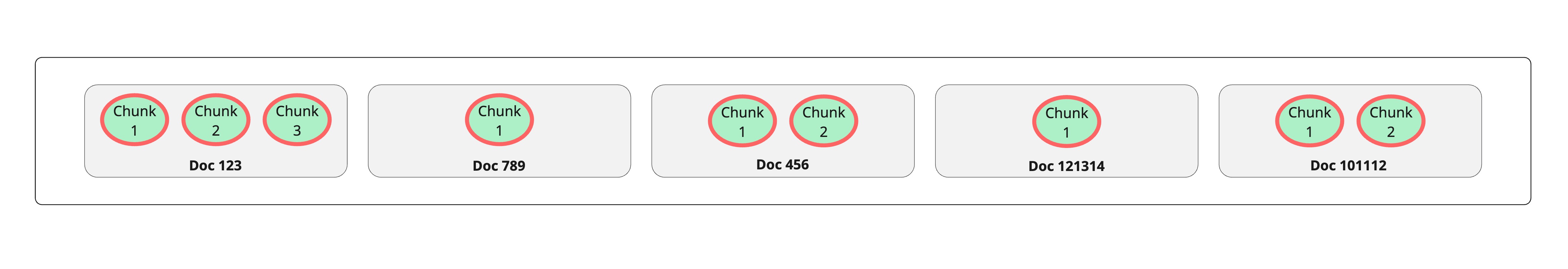}
    \label{fig:final_context_creation}
  \end{subfigure}

  \caption{Stage 2 of the Recall Pipeline: (a) Chunk extraction from query subgraph based on tags obtained from Stage 1 (b) Final Context}
  \label{fig:recall_pipeline_stage2}
\end{figure*}

\section{BambooKG: Architecture and Pipelines}

\subsection{Memorisation Pipeline}
The core concept of our approach lies in a multi-stage pipeline that constructs the knowledge graph. The overall pipeline is shown in Figure \ref{fig:memory_pipeline_all}. This pipeline consists of three primary stages: Chunking, Tag Generation, and Knowledge Graph Creation.

\subsubsection{\textbf{Chunking Stage}}

In the first stage, the input document is segmented into semantically coherent text blocks, chunks, following the standard text segmentation methods \cite{pak2018textsegmentation}. Each chunk is set at a fixed value that can range between 200–1200 tokens, depending on the overall document length (Figure \ref{fig:memory_pipeline_all}a). For instance, a document may be divided into multiple chunks, each capturing a coherent fragment of meaning (e.g., sentences or short paragraphs).

\subsubsection{\textbf{Tag Generation Stage}}

Each chunk is then passed to the Tagger, which is implemented as a controlled LLM call with a restricted prompt designed to extract key semantic entities or concepts, referred to as tags. The Tagger outputs a fixed-length list of tags per chunk, representing the most salient or contextually important terms identified by the model (Figure \ref{fig:memory_pipeline_all}b).

The overall quality and interpretability of the downstream knowledge graph depend critically on what the Tagger determines to be “important.” In this study, the model was intentionally left general to evaluate baseline performance; however, domain-specific prompting can be used to bias tag selection toward certain concepts and suppress less relevant ones, thereby controlling the signal-to-noise ratio of the resulting graph (see Section \ref{sec:future_work}).

Unlike conventional triplet-based knowledge graph extraction methods \cite{etzioni2015openie}, which rely on rigid subject-predicate-object relationships, our tag-based approach is structurally flexible. This flexibility offers two key advantages:

It eliminates the constraint of directional, grammar-bound triplets, enabling the capture of co-occurring concepts that may not conform to syntactic relations.

It facilitates the introduction of constrained tag vocabularies (domain-specific or general) in future iterations, providing a tunable framework for graph refinement and denoising.

\subsubsection{\textbf{Knowledge Graph Creation Stage}}

After tag generation, a BambooKG subgraph is constructed for each chunk and incrementally merged into the global BambooKG. Each tag is represented as a node, while edges denote co-occurrence relationships between tags within the same chunk (Figure \ref{fig:memory_pipeline_all}c)

Edge weights correspond to the frequency of co-occurrence - that is, the number of chunks in which a given tag pair appears together. This frequency-weighted graph structure reflects an undirected associative network, where stronger weights indicate tighter semantic coupling between tags.
Conceptually, this process parallels Spike-Timing Dependent Plasticity (STDP) in neuroscience, where connections between neurons that “fire together” are strengthened, forming the basis of associative memory \cite{caporale2008stdp}. Similarly, in BambooKG, each update to edge weights signifies that the system has “memorised” new knowledge.

Additionally, a second knowledge graph is generated to map the tags to their originating chunks and documents for final context retrieval in the query pipeline.

In this work, we focus on establishing the baseline BambooKG without performing any explicit clustering, pruning, or noise reduction of tags. These post-processing and optimisation steps remain open directions for future exploration.

\subsection{Recall Pipeline}
During memory recall, the retrieval process operates over the existing BambooKG structure generated during memorisation. The process begins when a user submits a query, which is first passed through the Tagger. Unlike the memorisation phase, the Tagger’s vocabulary is now restricted to the global tag vocabulary of BambooKG (that is, only the tags which are contained within BambooKG), ensuring that only known concepts are recognised. If the Tagger fails to identify any valid tags, it is assumed that BambooKG has not yet “learned” the requested concept.

\subsubsection{\textbf{Query Tag Extraction}}

The Tagger extracts the set of query tags from the query (e.g., in “What pet eats fish?”, the tags pet and fish are identified from the existing vocabulary). These tags serve as starting points of search within the tag frequency-weight knowledge graph.

\subsubsection{\textbf{Subgraph Retrieval}}

For each query tag, BambooKG extracts a local subgraph to approximate the region of semantic relevance. Following a decay-based neighbourhood exploration, the top X first-degree neighbours (directly connected tags) and top Y second-degree neighbours (tags connected through an intermediary) are selected, ranked by their edge weights - i.e., the frequency of co-occurrence. An example of this process in shown in Figure \ref{fig:query_subgraph_generation} with X and Y being set at 2.

This selective neighbourhood expansion approximates a form of associative recall, where strongly connected concepts are prioritised over weak or peripheral associations. For clarity of visualisation, only the top-ranked connections are displayed in the diagram, though the full graph traversal may include additional edges.

\subsubsection{\textbf{Context Construction}}

Once the subgraph is defined, BambooKG identifies all document chunks that contributed to the formation of the retrieved edges (Figure \ref{fig:recall_pipeline_stage2}a). These chunks represent the episodic context associated with the concepts recalled, analogous to the reactivation of cortical traces by the hippocampus during biological memory recall \cite{Wiltgen2010HippocampusSelectiveRetrieval}. In this sense, BambooKG functions as a synthetic hippocampal index, reactivating distributed memory fragments that co-occurred with the query tags during learning.

The retrieved chunks, aggregated across all relevant documents, form the final context. This context is then provided to the LLM, which synthesises an answer grounded in the recalled knowledge (Figure \ref{fig:recall_pipeline_stage2}b).

Importantly, BambooKG supports partial pattern matching, allowing recall to succeed even when the full combination of tags has not been observed before. If a query partially overlaps with known subgraphs (e.g., pet is known but fish is new), the system can still infer context from related neighbours (e.g., cat, dog, indoor) and construct an approximate answer. This mechanism parallels pattern completion observed in hippocampal memory systems, enabling the model to generalise from incomplete cues \cite{Grande2019HolisticRecollection}.

\begin{table*}[t]
\caption{Results on HotPotQA dataset.}
\label{tab:hotpotqa}
\vskip 0.15in
\begin{center}
\begin{small}
\begin{sc}
\begin{tabular}{lcccc}
\toprule
Method & Top-K & Accuracy (\%) & Avg. Context Size (Tokens) & Avg. Retrieval Time (s) \\
\midrule
RAG       & 5 & 71 & 648 & 2.16 \\
OpenIE    & 5-3      & 57         & 264             & 4.55 \\
GraphRAG  & N/A      & 20         & N/A         & 4.98 \\
KGGen     & 5-3      & 71         & 440             & 3.45 \\
BambooKG  & 5-3      & 78         & 1,887           & 0.01 \\
\bottomrule
\end{tabular}
\end{sc}
\end{small}
\end{center}
\vskip -0.1in
\end{table*}

\begin{table*}[t]
\caption{Results on MuSiQue dataset.}
\label{tab:musique}
\vskip 0.15in
\begin{center}
\begin{small}
\begin{sc}
\begin{tabular}{ccccc c}
\toprule
Hops & Method & Top-K & Accuracy (\%) & Avg. Context Size (Tokens) & Avg. Retrieval Time (s) \\
\midrule
2 & RAG       &  5 & 58 & 525 & 5.65 \\
  & OpenIE    & 5-3      & 20         & 275             & 4.96 \\
  & GraphRAG  & N/A      & 45         & N/A         & 6.05 \\
  & KGGen     & 5-3      & 41         & 295             & 3.11 \\
  & BambooKG  & 5-3      & 69         & 2,905           & 0.01 \\
\midrule
3 & RAG       & 5 & 14 & 1,078 & 6.12 \\
  & OpenIE    & 5-3      & 1          & 280               & 2.51 \\
  & GraphRAG  & N/A      & 33         & N/A           & 8.90 \\
  & KGGen     & 5-3      & 10         & 318               & 2.43 \\
  & BambooKG  & 5-3      & 54         & 16,273            & 0.01 \\
\midrule
4 & RAG       & 5 & 53 & 748 & 5.59 \\
  & OpenIE    & 5-3      & 6          & 328               & 2.93 \\
  & GraphRAG  & N/A      & 51         & N/A           & 8.21 \\
  & KGGen     & 5-3      & 8          & 228               & 2.23 \\
  & BambooKG  & 5-3      & 56         & 11,728            & 0.01 \\
\midrule
\multicolumn{6}{l}{\textbf{Average}} \\
\midrule
\multicolumn{2}{c}{RAG}      & 5 & 42 & 784 & 5.79 \\
\multicolumn{2}{c}{OpenIE}   & 5-3      & 9          & 294             & 3.47 \\
\multicolumn{2}{c}{GraphRAG} & N/A      & 43         & N/A         & 7.72 \\
\multicolumn{2}{c}{KGGen}    & 5-3      & 20         & 280             & 2.59 \\
\multicolumn{2}{c}{BambooKG} & 5-3      & 60         & 10,301          & 0.01 \\
\bottomrule
\end{tabular}
\end{sc}
\end{small}
\end{center}
\vskip -0.1in
\end{table*}

\section{Experimental Setup}
\label{sec:experimental_setup}
The ability of an LLM to effectively answer a given question depends on the ability of the recall pipeline to accurately retrieve the relevant information.

In this study, we used accuracy as a primary measure of knowledge graph performance. We compared BambooKG with three other knowledge graph methods: OpenIE, GraphRAG and KGGen \cite{luo2025causalgraphrag, etzioni2015openie, MoYuKazdan2025_KGGen}, as well as standard RAG serving as a baseline benchmark. As the other knowledge graph approaches are designed for embedding-based search algorithms, instead of using weighted edges to select top-k chunks, we used top-k embeddings as in ordinary RAG. The top-k for BambooKG, OpenIE and KGGen was set at 5-3 (5 first-degree neighbours and 3 second-degree neighbours), whilst for RAG the top-k was chosen to be 5. GraphRAG did not have an ability to choose the top-k chunks to be retrieved \cite{edge2024graphrag}.

We chose two datasets to evaluate the performance of BambooKG: HotPotQA \cite{yang2018hotpotqa} for general knowledge recall and MuSiQue, which demands multi-hop knowledge retention and navigation and is considered to be one of the most challenging datasets of this kind \cite{trivedi2021musique}.

For the purposes of cost limitation, we randomly selected 100 questions (both correct and distractor) from the HotPotQA dataset and 100 questions for each of the 2, 3 and 4 hops from the MuSiQue dataset.

To evaluate accuracy, we sent the recalled subgraph to GPT-4o to generate an answer, and then used the same model as LLM-as-a-Judge \cite{Gu2025SurveyLLMAsJudge} to evaluate whether the predicted answer matches the expected answer. It is important to note that since GPT-4o is a non-deterministic model, the results will vary slightly on each run \cite{Song2024GoodBadGreedy}.

\section{Results}

We used 100 randomly selected questions from HotPotQA and MuSiQue datasets to benchmark BambooKG. Additionally, in each dataset, we compared BambooKG against simple RAG, OpenIE, GraphRAG and KGGen. The full results are presented in Tables \ref{tab:hotpotqa} and \ref{tab:musique}.

Using these benchmarks, we demonstrate that BambooKG outperforms other knowledge graph methods in two critical parameters: data recall and average retrieval time, in both simple questions (HotPotQA) and multi-hop questions (MuSiQue).

We believe that there are several reasons for this performance. To start with minor issues, OpenIE often generates incoherent or nonsensical triplets (“if" as a valid node). GraphRAG generates a small number of nodes per article, resulting in information being missed. KGGen, in turn, performs well on simple questions (HotpotQA) but struggles with multi-hop questions due to poorly performing clusters.

Curiously, GraphRAG showed inferior performance in HotPotQA in spite of its relatively good stance as shown in other research \cite{xu2025noderag}. We believe the reason for this is twofold: GraphRAG KG tends to miss answer node entities \cite{Han2025RAGvsGraphRAG}; and that the corpus contained distractor documents. We have used both the correct documents (supporting facts) and distractor documents taken from the context field to generate the corpus for the 100 randomly selected questions. This most likely led to incorrect community generation, which hindered the performance of GraphRAG.

In addition to the weighted edges, two seemingly counterintuitive factors contribute to the satisfactory performance of BambooKG: lack of triplets and the usage of arbitrary nodes. At the expense of the increased graph size and the loss of rigid structure, this leads to the decreased information loss whilst preserving epistemic connectivity between separate documents. Additionally, the difficulty of forming adequate single- or few-word embeddings when applying RAG to knowledge graph triplets means that embedding-based search approaches would naturally experience information loss, as well as increased extraction time \cite{Amur2023STSS, edge2024graphrag}.

Another point of note is that BambooKG pipelines only use a single LLM call: one during the Memorisation Pipeline, to trigger the Tagger. The Recall Pipeline proceeds entirely without LLMs or embeddings, resulting in highly competitive extraction speeds.

One downside of BambooKG is the increased context size when compared to other approaches. However, this is beyond the scope of this work, as the context window is entirely dependent on the LLM one uses and is wholly unrelated to the long-term memory approaches.

\section{Future Work}
\label{sec:future_work}

The fundamental limitation of the BambooKG performance lies in the way that Tagger works. For the experiments run for this paper, the Tagger chooses arbitrary, generic tags, effectively summarising the text chunk into a set of discrete words. By focusing the Tagger's attention on a specific domain (via fine-tuning or prompt engineering), we can achieve greater data retention and recall rates on specialised corpora. Another point of future research is the organic formation of communities and clusters (with or without LLM calls), which will become critical for graph navigation when the volume of information added into the knowledge graph becomes significantly large. Additionally, the selection and extraction of a subgraph during the recall stage demands further refinement to bring the context size down and accelerate the final, LLM-based decision-making.





\bibliography{example_paper}
\bibliographystyle{mlsys2025}

\end{document}